# Diagnostic Reasoning Prompts Reveal the Potential for Large Language Model Interpretability in Medicine

Thomas Savage MD[1,2], Ashwin Nayak MD MS[1,2], Robert Gallo MD[3], Ekanath Rangan MD[1], Jonathan H Chen MD PhD[1,2,4,5]

[1] Department of Medicine, Stanford University, Stanford, California
[2] Division of Hospital Medicine, Stanford University, Stanford, CA
[3] Palo Alto Veterans Affairs Medical Center
[4] Stanford Center for Biomedical Informatics Research, Stanford University, Stanford, CA
[5] Clinical Excellence Research Center, Stanford University, Stanford, CA

## Abstract

One of the major barriers to using large language models (LLMs) in medicine is the perception they use uninterpretable methods to make clinical decisions that are inherently different from the cognitive processes of clinicians. In this manuscript we develop novel diagnostic reasoning prompts to study whether LLMs can perform clinical reasoning to accurately form a diagnosis. We find that GPT-4 can be prompted to mimic the common clinical reasoning processes of clinicians without sacrificing diagnostic accuracy. This is significant because an LLM that can use clinical reasoning to provide an interpretable rationale offers physicians a means to evaluate whether LLMs can be trusted for patient care. Novel prompting methods have the potential to expose the "black box" of LLMs, bringing them one step closer to safe and effective use in medicine.

Large Language Models (LLMs) have received widespread attention for their human-like performance on a wide variety of text-based tasks. Within medicine, initial efforts have demonstrated that LLMs can write clinical notes[1], pass standardized medical exams[2], and draft responses to patient questions.[3,4] In order to integrate LLMs more directly into clinical care, it is imperative to better understand their clinical reasoning capabilities.

Clinical reasoning is a set of problem-solving processes specifically designed for diagnosis and management of a patient's medical condition. Commonly used diagnostic techniques include differential diagnosis formation, intuitive reasoning, analytical reasoning, and Bayesian inference. Early assessments of the clinical reasoning abilities of LLMs have been limited, studying model responses to multiple-choice questions.[5–9] More recent work has focused on free-response clinical questions and suggests that newer LLMs, such as GPT-4, show promise in diagnosis of challenging clinical cases.[10,11]

Prompt engineering is emerging as a discipline in response to the phenomena that LLMs can perform substantially differently depending on how questions and prompts are posed to them.[12,13] Advanced prompting techniques have demonstrated improved performance on a range of tasks[14], while also providing insight into how LLMs came to a conclusion. A notable example is Chain-of-thought (CoT) prompting, which involves instructing the LLM to divide its task into smaller reasoning steps and then complete the task step-by-step.[15] Given that clinical reasoning tasks regularly use step-by-step processes, CoT prompts modified to reflect the cognitive processes taught to and utilized by clinicians might elicit better understanding of LLM performance on clinical reasoning tasks.

In this paper we evaluate the performance of GPT-3.5 and GPT-4[16] on open-ended clinical questions assessing diagnostic reasoning. Specifically, we evaluate LLM performance on a modified MedQA USMLE (United States Medical Licensing Exam) dataset[17]. We compare traditional CoT prompting with several novel "diagnostic reasoning" prompts that are modeled after the cognitive processes of differential diagnosis formation, intuitive reasoning, analytical reasoning, and Bayesian inference. This study assesses whether LLMs can demonstrate clinical reasoning abilities using specialized instructional prompts that combine clinical expertise and advanced prompting methods.

A modified version of the MedQA USMLE question dataset was used for this study. Questions were converted to free response by removing the multiple-choice options after the question stem. Only Step 2 and Step 3 USMLE questions were included, as Step 1 questions focus heavily on memorization of facts rather than clinical reasoning skills.[10] Only questions evaluating the task of diagnosing a patient were included to simplify prompt engineering. A training set of 95 questions was used for iterative prompt development and a test set of 518 questions was reserved for evaluation. The full test set can be found in Supplemental Information I.

One traditional CoT prompt and 4 clinical reasoning prompts were developed (differential diagnosis, analytical, Bayesian and intuitive reasoning). Each prompt included two example questions (Table 1) with rationales employing the target reasoning strategy. This is a technique known as few-shot learning.[12] The full prompts used for the MedQA dataset are provided in Table 2.

---

**Example Question 1**
Shortly after undergoing a bipolar prosthesis for a displaced femoral neck fracture of the left hip acquired after a fall the day before, an 80-year-old woman suddenly develops dyspnea. The surgery under general anesthesia with sevoflurane was uneventful, lasting 98 minutes, during which the patient maintained oxygen saturation readings of 100% on 8 L of oxygen. She has a history of hypertension, osteoporosis, and osteoarthritis of her right knee. Her medications include ramipril, naproxen, ranitidine, and a multivitamin. She appears cyanotic, drowsy, and is oriented only to person. Her temperature is 38.6°C (101.5°F), pulse is 135/minute, respirations are 36/min, and blood pressure is 155/95 mm Hg. Pulse oximetry on room air shows an oxygen saturation of 81%. There are several scattered petechiae on the anterior chest wall. Laboratory studies show a hemoglobin concentration of 10.5 g/dL, a leukocyte count of 9,000/mm3, a platelet count of 145,000/mm3, and a creatine kinase of 190 U/L. An ECG shows sinus tachycardia. What is the most likely diagnosis?

---

**Example Question 2**
A 55-year-old man comes to the emergency department because of a dry cough and severe chest pain beginning that morning. Two months ago, he was diagnosed with inferior wall myocardial infarction and was treated with stent implantation of the right coronary artery. He has a history of hypertension and hypercholesterolemia. His medications include aspirin, clopidogrel, atorvastatin, and enalapril. His temperature is 38.5°C (101.3°F), pulse is 92/min, respirations are 22/min, and blood pressure is 130/80 mm Hg. Cardiac examination shows a high-pitched scratching sound best heard while sitting upright and during expiration. The remainder of the examination shows no abnormalities. An ECG shows diffuse ST elevations. Serum studies show a troponin I of 0.005 ng/mL (N < 0.01). What is the most likely cause of this patient's symptoms?

---

Table 1. Example questions used in all MEDQA prompts provided in Table 2.

| | Traditional CoT Reasoning Prompt | |
|---|---|---|
| Traditional Chain-of-Thought (CoT) | **Prompt:**<br>*Provide a step-by-step deduction that identifies the correct response*<br><br>**{Example Question 1}**<br>**Example Rationale 1:**<br>The patient had a surgical repair of a displaced femoral neck fracture. The patient has petechiae. The patient has a new oxygen requirement, meaning they are having difficulty with their breathing. This patient most likely has a fat embolism.<br>**{Example Question 2}**<br>**Example Rationale 2:**<br>This patient is having chest pain. They recently had a heart attack and has new chest pain, suggesting he may have a problem with his heart. The EKG has diffuse ST elevations and he has a scratching murmur. This patient likely has Dressler Syndrome. | |
| | **Clinical Reasoning Prompts** | |
| Differential Diagnosis CoT | **Prompt:**<br>*Use step by step deduction to create a differential diagnosis and then use step by step deduction to determine the correct response.*<br><br>**{Example Question 1}**<br>**Example Rationale 1:**<br>This patient has shortness of breath after a long bone surgery. The differential for this patient is pulmonary embolism, fat embolism, myocardial infarction, blood loss, anaphylaxis, or a drug reaction. The patient has petechiae which makes fat embolism more likely. This patient most likely has a fat embolism.<br>**{Example Question 2}**<br>**Example Rationale 2:**<br>This patient has chest pain with diffuse ST elevations after a recent myocardial infarction. The differential for this patient includes: myocardial infarction, pulmonary embolism, pericarditis, Dressler syndrome, aortic dissection, and costochondritis. This patient likely has a high-pitched scratching sound on auscultation associated with pericarditis and Dressler Syndrome. This patient has diffuse ST elevations associated with Dressler Syndrome. This patient most likely has Dressler Syndrome. | |
| Intuitive Reasoning CoT | **Prompt:**<br>*Use symptom, signs, and laboratory disease associations to step by step deduce the correct response.*<br><br>**{Example Question 1}**<br>**Example Rationale 1:**<br>This patient has findings of petechiae, altered mental status, shortness of breath, and recent surgery suggesting a diagnosis of fat emboli. The patient most likely has a fat embolism.<br>**{Example Question 2}**<br>**Example Rationale 2:**<br>This patient had a recent myocardial infarction with new development of diffuse ST elevations, chest pain, and a high pitched scratching murmur which are found in Dressler's syndrome. This patient likely has Dressler's Syndrome. | |
| Analytic Reasoning CoT | **Prompt:**<br>*Use analytic reasoning to deduce the physiologic or biochemical pathophysiology of the patient and step by step identify the correct response.*<br><br>**{Example Question 1}**<br>**Example Rationale 1:**<br>The patient recently had large bone surgery making fat emboli a potential cause because the bone marrow was manipulated. Petechiae can form in response to capillary inflammation caused by fat emboli. Fat micro globules cause CNS microcirculation occlusion causing confusion and altered mental status. Fat obstruction in the pulmonary arteries can cause tachycardia and shortness of breath as seen in this patient. This patient most likely has a fat embolism.<br>**{Example Question 2}**<br>**Example Rationale 2:**<br>This patient had a recent myocardial infarction which can cause myocardial inflammation that causes pericarditis and Dressler Syndrome. The diffuse ST elevations and high pitched scratching murmur can be signs of pericardial inflammation as the inflamed pericardium rubs against the pleura as seen with Dressler Syndrome. This patient likely has Dressler Syndrome. | |
| Bayesian Reasoning CoT | **Prompt:**<br>*Use step-by-step Bayesian Inference to create a prior probability that is updated with new information in the history to produce a posterior probability and determine the final diagnosis.*<br><br>**{Example Question 1}**<br>**Example Rationale 1:**<br>The prior probability of fat embolism is 0.05% however the patient has petechiae on exam which is seen with fat emboli, which increases the posterior probability of fat embolism to 5%. Altered mental status increases the probability further to 10%. Recent orthopedic surgery increases the probability of fat emboli syndrome to 60%. This patient most likely has a fat embolism.<br>**{Example Question 2}**<br>**Example Rationale 2:**<br>The prior probability of Dressler Syndrome is 0.01%. The patient has diffuse ST elevations, increasing the probability of Dressler Syndrome to 5%. The patient has a scratching murmur which increases the probability to 10%. In the setting of a recent MI the posterior probability of myocardial infarction is 55%. This patient likely has Dressler Syndrome. | |

Table 2. The traditional and diagnostic CoT prompts evaluated in this study. The provided rationales are in response to the example questions provided in Table 1. Note that the full text of the example questions has been omitted for readability.

Example LLM responses for each prompting strategy can be found in Figure 1 for GPT-3.5 and Figure 2 for GPT-4. Full results can be found in Supplemental Information I.

| |
|---|
| **Question :** A 30-year-old African American woman comes to the physician for the evaluation of a dry cough and chest discomfort for the past 3 days. During this period, the patient has had headaches, muscle aches, joint pain, fever, and chills. Ten days ago, she was hiking with her family in Mississippi. The patient has asthma that is treated with an albuterol inhaler. Her mother has a lung disease treated with methotrexate. The patient has smoked one pack of cigarettes daily for the past 10 years. Her temperature is 38°C (100.4°F). Physical examination shows slight wheezes throughout both lung fields. Laboratory studies and urinalysis are positive for polysaccharide antigen. Bronchoalveolar lavage using silver/PAS-staining shows macrophages filled with a dimorphic fungus with septate hyphae. What is the most likely cause of this patient's symptoms?<br><br>Answer: Histoplasmosis |
| **Traditional Chain-of-Thought (GPT-3.5)**<br>*'The patient has a dry cough, chest discomfort, and other symptoms such as fever, chills, headaches, and muscle aches. She has a positive polysaccharide antigen and bronchoalveolar lavage shows a dimorphic fungus with septate hyphae. This patient likely has Valley Fever.'* |
| **GPT-3.5 Differential Diagnosis CoT (GPT-3.5)**<br>*'The differential diagnosis for this patient includes: pneumonia, tuberculosis, eosinophilic pneumonia, hypersensitivity pneumonitis, and coccidioidomycosis. The patient's recent history of hiking in Mississippi and the presence of a dimorphic fungus with septate hyphae on bronchoalveolar lavage make coccidioidomycosis the most likely cause of her symptoms.'* |
| **GPT-3.5 Intuitive Reasoning CoT (GPT-3.5)**<br>*'The patient has a history of asthma, which increases the likelihood of asthma exacerbation. The positive result for polysaccharide antigen suggests atypical pneumonia. The presence of a dimorphic fungus with septate hyphae suggests the presence of fungal pneumonia. This patient most likely has fungal pneumonia.'* |
| **Analytic Reasoning CoT (GPT-3.5)**<br>*'The patient's recent history of hiking in Mississippi and the positive polysaccharide antigen and silver/PAS-staining results strongly suggest fungal pneumonia as the cause of her symptoms.'* |
| **Bayesian Inference CoT (GPT-3.5)**<br>*'The prior probability of blastomycosis is 1%, however the patient's recent travel increases the probability to 5%. The laboratory and urinalysis results increase the posterior probability to 30%. The presence of a dimorphic fungus on bronchoalveolar lavage increases the probability to 60%. This patient is likely suffering from blastomycosis.'* |

Figure 1. Example GPT-3.5 rationales responding to a traditional CoT prompt as well as diagnostic reasoning prompts. LLM response and rationale results for the entire test set can be found in supplemental information I.

| Chain-of-Thought (GPT-4) |
|---|
| 'This patient has recently been hiking in Mississippi, presenting with symptoms of fever, chills, dry cough, and chest discomfort. She also has a history of asthma and a heavy smoking habit, which might predispose her to respiratory infections. She tests positive for a polysaccharide antigen and bronchoalveolar lavage reveals a dimorphic fungus with septate hyphae in macrophages. The dimorphic nature of the fungus and the septate hyphae are characteristic of Histoplasma capsulatum, a fungus endemic to the Mississippi River Valley. Therefore, the most likely cause of this patient's symptoms is histoplasmosis, a fungal infection caused by inhalation of Histoplasma capsulatum spores.' |

| Differential Diagnosis Rationale |
|---|
| 'The differential diagnosis in this case includes: community-acquired pneumonia (e.g., Streptococcus pneumoniae, Mycoplasma pneumoniae), allergic bronchopulmonary aspergillosis (given her asthma history and smoking habit), tuberculosis (given her mother's methotrexate use), and histoplasmosis (given her travel history to Mississippi and finding of dimorphic fungi).

*Looking at the given information:*

She was hiking in Mississippi 10 days ago - an endemic area for histoplasmosis. Histoplasma capsulatum is a dimorphic fungus, meaning it can exist in two forms. In the environment, it exists as a mold with septate hyphae. In the body, it converts to a yeast form. It resides within macrophages after inhalation, which corresponds with the bronchoalveolar lavage findings.
Her symptoms of fever, chills, dry cough, chest discomfort, headaches, muscle aches, and joint pain, are consistent with acute histoplasmosis. Additionally, polysaccharide antigen tests can be used to detect Histoplasma capsulatum in urine, blood, or other body fluids, which supports the diagnosis.
While allergic bronchopulmonary aspergillosis could theoretically present with similar respiratory symptoms, it doesn't align with the presentation of a dimorphic fungus with septate hyphae residing within macrophages.
Thus, taking all this into consideration, the most likely diagnosis for her symptoms is histoplasmosis.' |

| Intuitive Reasoning |
|---|
| 'This patient presents with fever, cough, chest discomfort, and systemic symptoms such as headaches, muscle aches, and joint pain after hiking in Mississippi. The detection of a polysaccharide antigen in laboratory studies and a dimorphic fungus with septate hyphae in bronchoalveolar lavage suggest a fungal infection, most likely Histoplasma capsulatum. This fungus is endemic in the Mississippi River valley and can cause a systemic infection with pulmonary and extrapulmonary manifestations in humans, especially in those with chronic respiratory conditions like asthma. These factors together suggest that this patient's symptoms are likely due to Histoplasmosis.' |

| Analytic Reasoning |
|---|
| 'The patient's history of recent hiking in Mississippi combined with her symptoms of dry cough, chest discomfort, headaches, muscle aches, joint pain, fever, and chills suggests an infectious process, specifically a fungal infection. The detection of polysaccharide antigen and the bronchoalveolar lavage findings of macrophages filled with a dimorphic fungus with septate hyphae are key to diagnosing the cause. The fungus described is consistent with Histoplasma capsulatum, which is endemic in the Mississippi River Valley area. The mold form of Histoplasma, found in soil, transforms into the yeast form in human body temperature and can lead to a systemic disease known as Histoplasmosis. The yeast form is engulfed by macrophages, as seen in the bronchoalveolar lavage, and can cause the symptoms presented by the patient.' |

| Bayesian Inference |
|---|
| 'The prior probability of Histoplasmosis (a fungal infection often acquired through inhalation of spores in certain geographic areas such as the Mississippi River Valley) given the patient's recent hiking trip in Mississippi could be assumed at around 10%. The presence of fever, cough, and chest discomfort increases the likelihood of this being a respiratory infection, raising the probability to perhaps 20%. Further, the fact that the patient's mother is being treated with methotrexate suggests possible genetic predisposition to immune-related diseases, although it doesn't directly imply histoplasmosis, it can increase the suspicion to around 25%.

The patient's condition has not improved with the use of an albuterol inhaler, which would typically ameliorate symptoms of asthma, this suggests an alternative cause for the respiratory symptoms, perhaps raising the probability of Histoplasmosis to 30%. The significant detail here is the identification of a dimorphic fungus with septate hyphae in a bronchoalveolar lavage sample. Histoplasma is a dimorphic fungus, which substantially increases the posterior probability of Histoplasmosis to above 80% or even higher.' |

Figure 2. Example GPT-4 rationales responding to the question posed in Figure 1. LLM response and rationale results for the entire test set can be found in Supplemental Information I.

GPT-3.5 correctly answered 46% of questions using traditional CoT prompting. Among the clinical reasoning prompts, GPT-3.5 achieved the highest performance with intuitive reasoning (48% vs 46%, difference of 1.7%, CI -1.6% to +5.1%, P= 0.9). Compared to traditional CoT, GPT-3.5's performance was significantly worse with analytic reasoning (40%, difference of -6%, CI -10% to -2.5%, P= 0.001), differential diagnosis formation (38%, difference of -8%, CI -13% to -4.5%, P= < 0.001), and Bayesian inference (42%, difference of -4%, CI -8.1% to -0.7%, P = 0.02). Inter-rater agreement for the MedQA GPT-3.5 evaluation was 97% with a Cohen's Kappa of 0.93.

The GPT-4 API generated an error for 21 questions of the test set, reducing the test set size to 497. Overall, GPT-4 demonstrated improved accuracy over GPT-3.5. GPT-4 achieved an accuracy of 76% with traditional CoT, 77% with intuitive reasoning (+0.8%, CI -2.7% to +4.3%, P= 0.73), 78% with differential diagnosis (+2.2%, CI -1.4% to +5.8%, P= 0.24), 78% with analytic reasoning (+1.6%, CI -1.5% to +4.8%, P= 0.35), and 72% with Bayesian Inference (-3.4%, CI -7.0% to +0.3%, P= 0.07). Inter-rater agreement for the GPT-4 MedQA evaluation was 99% with a Cohen's Kappa of 0.98.

In this study we found that GPT-3.5 performance was similar with traditional and intuitive reasoning CoT prompts, but significantly worse with differential diagnosis, analytical, and Bayesian inference CoT prompts. These findings suggest GPT-3.5 is not able to use advanced clinical reasoning processes to arrive at an accurate diagnosis. In contrast, GPT-4 demonstrated similar performance between traditional and diagnostic reasoning CoT prompts, suggesting it can be prompted to successfully perform clinical reasoning processes without sacrificing diagnostic accuracy. Given that the same prompts were used for both models, these results highlight the significant advancement in reasoning abilities between GPT-3.5 and GPT-4.

The finding that GPT-4 can successfully imitate the same cognitive processes as physicians to arrive accurately at an answer is significant and will have implications for how language models can be used in the clinical workflow. A model that not only provides an accurate diagnosis but also produces a clinical reasoning rationale to support it is a major step towards interpretability and trust. Strategies that align model outputs in this way could help transition LLMs away from a purely "black box" model to one in which final model outputs are always accompanied by a robust, interpretable rationale grounded in clinical reasoning (Figure 3). This will offer clinicians a means to evaluate how a LLM arrived at an answer and assess for appropriate quality and logic. Ultimately this framework holds the potential to transform LLM systems into interpretable tools that can credibly be used in medicine. Future work should further assess whether LLM clinical reasoning rationales are sufficiently logical, accurate and devoid of hallucinations.

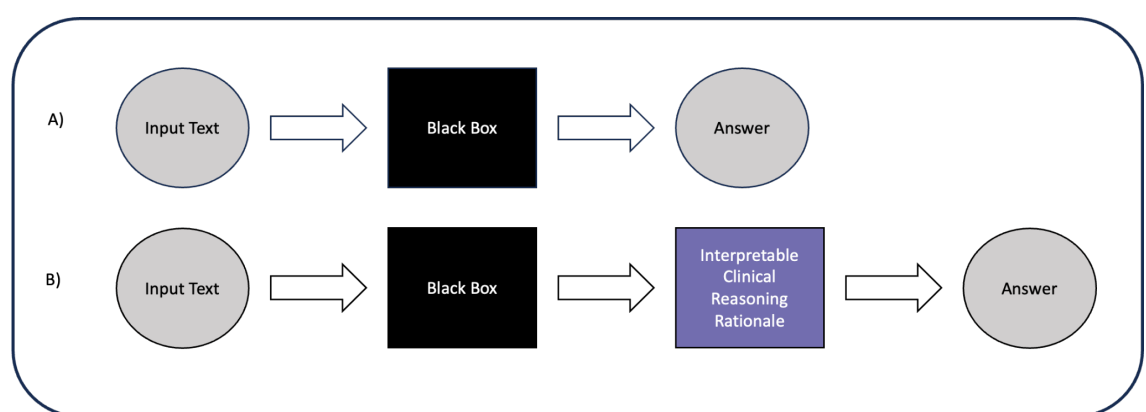

Figure 3. A) Current LLM workflow. B) Proposed LLM workflow.

The strengths of our investigation are a novel design that leverages chain-of-thought prompting for insight into LLM clinical reasoning capabilities as well as the use of free response clinical case questions where previous studies have been limited to multiple-choice or simple open-ended fact retrieval that do not challenge LLM clinical reasoning abilities. We designed our evaluation with free response questions both the USMLE to facilitate rigorous comparison between prompting strategies.

A limitation of our study is that while our prompt engineering process surveyed a wide range of prompt styles, we were not able to test all possible diagnostic reasoning CoT prompts. We hope that future studies can iterate on our diagnostic reasoning prompts and use our open dataset as a benchmark for additional evaluation.

## Methods

**LLM Prompt Development**
We used an iterative process known as prompt engineering to develop our diagnostic reasoning prompts. During this process, we experimented with several different types of prompts (Supplemental Information II). In each round of prompt engineering, we evaluated GPT-3.5 accuracy on the MEDQA training set (Supplemental Information III). We found prompts that encouraged step-by-step reasoning without specifying what the steps should be, yielded better performance. We also found that prompts that focused on a single diagnostic reasoning strategy provided better results than prompts that combined multiple strategies.

**LLM Response Evaluation**
Language model responses were evaluated by physician authors AN, ER, RG and TS. Each question was evaluated by two blinded physicians. If there was disagreement in the grade assigned, a third evaluator determined the final grade. Any response that was felt to be equally correct and specific, as compared to the provided answer, was marked as correct.

**LLM Programming and Computing Resources**
For this evaluation we used the OpenAI Davinci-003 model via an OpenAI API to provide GPT-3.5 responses and GPT-4 model via an OpenAI API to provide GPT-4 responses. Prompting of the GPT-3.5 model was performed with the Demonstrate-Search-Predict (DSP) Python module.[19] Self-consistency was applied to all GPT-3.5 Chain-of-Thought prompts.[20] GPT-4 responses did not use DSP or self-consistency because those features were not available for GPT-4 at the time of submission. Computing was performed in a Google CoLab Jupyter Notebook. Full code can be found in Supplemental Information IV.

**Statistical Evaluation**
Statistical significance and confidence intervals were calculated against traditional CoT using McNemar's test for paired proportions, two-tailed. Statistical significance was set at an alpha of 0.05. Inter-rater agreement was assessed using Cohen's Kappa Statistic.

# Supplemental Information

**Supplemental Information I**
See file "MedQA_complete_graded_data.csv" for our complete test question set and prompt responses with grades for the MedQA evaluation of GPT-3.5 and GPT-4.

**Supplemental Information II**
In the process of engineering our clinical reasoning prompts, we experimented with many different prompting strategies. All prompts tested are included below. The first prompt listed of every section is the final prompt that was selected. The number in parentheses is the number of questions the prompt answered correctly from the development set. Full results of our development set evaluation are found in Supplemental Information VII.

### Intuitive Reasoning
1. Use symptom, signs, and laboratory disease associations to step by step deduce the correct response. (39)
2. Use disease association to deduce the correct response. (38)
3. First list the top 6 diagnoses that answer the question. Then reference the question to find patient information (past medical history, symptoms, physical exam findings, lab tests, imaging) that are associated with any of the diagnoses in the differential. Answer the question based on the diagnoses that are most likely. (39)
4. Create a differential diagnosis and then use disease association to deduce the correct response. (39)
5. First list the top 6 diagnoses that answer the question. Then reference the question to find patient information or test results that are associated with any of the diagnoses in the differential. Answer with the diagnosis that is most likely. (39)
6. Follow the following steps: 1) list a broad differential of 6 diagnoses that answer the question. 2) reference the question to find patient information or test results that make certain diagnoses on the differential more likely. 3) Narrow the differential to 3 diagnoses based. 4) Again reference the question to find information that makes one diagnosis more likely. 6) Answer with the most likely diagnosis. (39)

### Analytic Reasoning
1. Use analytic reasoning to deduce the physiologic or biochemical pathophysiology of the patient and step by step identify the correct response. (40)
2. Use analytic reasoning to deduce the physiologic or biochemical pathophysiology of the patient and identify the correct response. (30)
3. Systematically reference each piece of patient information or test result in the prompt, explain if and how each piece of information supports one diagnosis. Answer with the most likely diagnosis. (30)
4. First list the top 6 diagnoses that answer the question. Then systematically reference each piece of patient information or test result in the prompt, explain if and how each piece of information supports one of the diagnoses on the differential. Answer with the most likely diagnosis. (33)
5. First list the top 6 diagnoses that answer the question. Then systematically reference each piece of information in the prompt (past medical history, symptoms, physical exam findings, lab tests, imaging), explain if and how each piece of information supports one of the diagnoses on the differential. Select the diagnosis most likely based on which diagnosis is most supported. (32)
6. Create a differential diagnosis, then use analytic reasoning to deduce the physiologic or biochemical pathophysiology of the patient and identify the correct response. (35)

### Bayesian Inference
1. Use step-by-step Bayesian Inference to create a prior probability that is updated with new information in the history to produce a posterior probability and determine the final diagnosis. (48)
2. Calculate a Bayes prior probability for the most likely diagnosis. Then reference the question to find all important patient information (past medical history, symptoms, physical exam findings, lab tests, imaging) that helps determine the diagnosis. For each piece of information, estimate the likelihood of the diagnosis being considered and calculate a posterior probability. Select the most likely diagnosis. (35)
3. First list the top 6 diagnoses that answer the question. Calculate a Bayes prior probability for each diagnosis. Then reference the question to find all important patient information or test results that helps determine the diagnosis. For each piece of information, estimate a likelihood of this information with each diagnosis being considered. Calculate a posterior probability for each diagnosis. Select the answer with the highest posterior probability. (39)
4. Create a differential diagnosis, then use a step-by-step Bayesian inference to deduce the correct response. (41)

### Differential Diagnosis
1. Use step by step deduction to create a differential diagnosis and then use step by step deduction to determine the correct response. (33)
2. First list the top 6 diagnoses that answer the question, select the diagnosis that is most likely. (30)
3. Create a differential diagnosis and use step by step deduction to determine the correct response. (33)
4. Follow the following steps: 1) list a broad differential of 6 diagnoses that answer the question. 2) reference the question to find patient information or test results that make certain diagnoses on the differential more likely. 3) Narrow the differential to 3 diagnoses based. 4) Again reference the question to find information that makes one diagnosis more likely. 6) Answer with the most likely diagnosis. (33)

**Supplemental Information III**
See file "Dev_Results.csv" for complete results of the Supplemental Information VI prompts on our Dev set.

**Supplemental Information IV**
See file "GPT3_API.ipynb" for our complete code for submission of prompts to the GPT-Davinici-003 API.

See file "GPT4_API.ipynb" for our complete code for submission of prompts to the GPT-4 API.